\documentclass[a4paper, 10pt, conference]{ieeeconf}      

\IEEEoverridecommandlockouts                              

\usepackage{color}
\usepackage[pdftex]{graphicx}

\usepackage{mathptmx} 
\usepackage{times} 
\usepackage{amsmath} 
\usepackage{amssymb}  

\usepackage[noadjust,nospace]{cite}

\usepackage{algorithm}
\usepackage{algpseudocode}
\usepackage{array}
\usepackage[caption=false,font=footnotesize]{subfig}
\usepackage{url}

\usepackage{enumerate}

\graphicspath{
  {figs/}
}

\title{\LARGE \bf
  A Real-Time Deep Learning Pedestrian Detector for Robot Navigation
}

 \author{David Ribeiro, Andr\'{e} Mateus, Pedro Miraldo, and Jacinto C. Nascimento
 \\ {Instituto de Sistemas e Rob\'{o}tica, Instituto Superior T\'{e}cnico, Universidade de Lisboa, Lisboa, Portugal}
 \thanks{978-1-5090-6234-8/17/\$31.00~\copyright{}2017 IEEE \hfill} 
 }%

\begin{document}
\maketitle
\thispagestyle{empty}
\pagestyle{empty}

\begin{abstract}
  A real-time Deep Learning based method for Pedestrian Detection (PD) is applied to the Human-Aware robot navigation problem. The pedestrian detector combines the Aggregate Channel Features (ACF) detector with a deep Convolutional Neural Network (CNN) in order to obtain fast and accurate performance. Our solution is firstly evaluated using a set of real images taken from onboard and offboard cameras and, then, it is validated in a typical robot navigation environment with pedestrians (two distinct experiments are conducted). The results on both tests show that our pedestrian detector is robust and fast enough to be used on robot navigation applications.
\end{abstract}

\section{Introduction}
In the last few years, robotics has become focused on Human-Robot Interaction and on its role in social environments. Some types of interaction can be, for example: speech; object handover; or a simple navigation behavior, where the robot needs to know if some obstacles are people or not, to decide what is the correct behavior. The study of robot navigation in the presence of people is called Human-Aware Navigation (HAN). For any type of Human-Robot Interaction, one needs to know the position of the people in these environments. Therefore, the Pedestrian Detection (PD) method is one of the most important steps for the robot to interact correctly with the humans.

\begin{figure}[t]
  \begin{center}
    \includegraphics[width=0.48\textwidth]{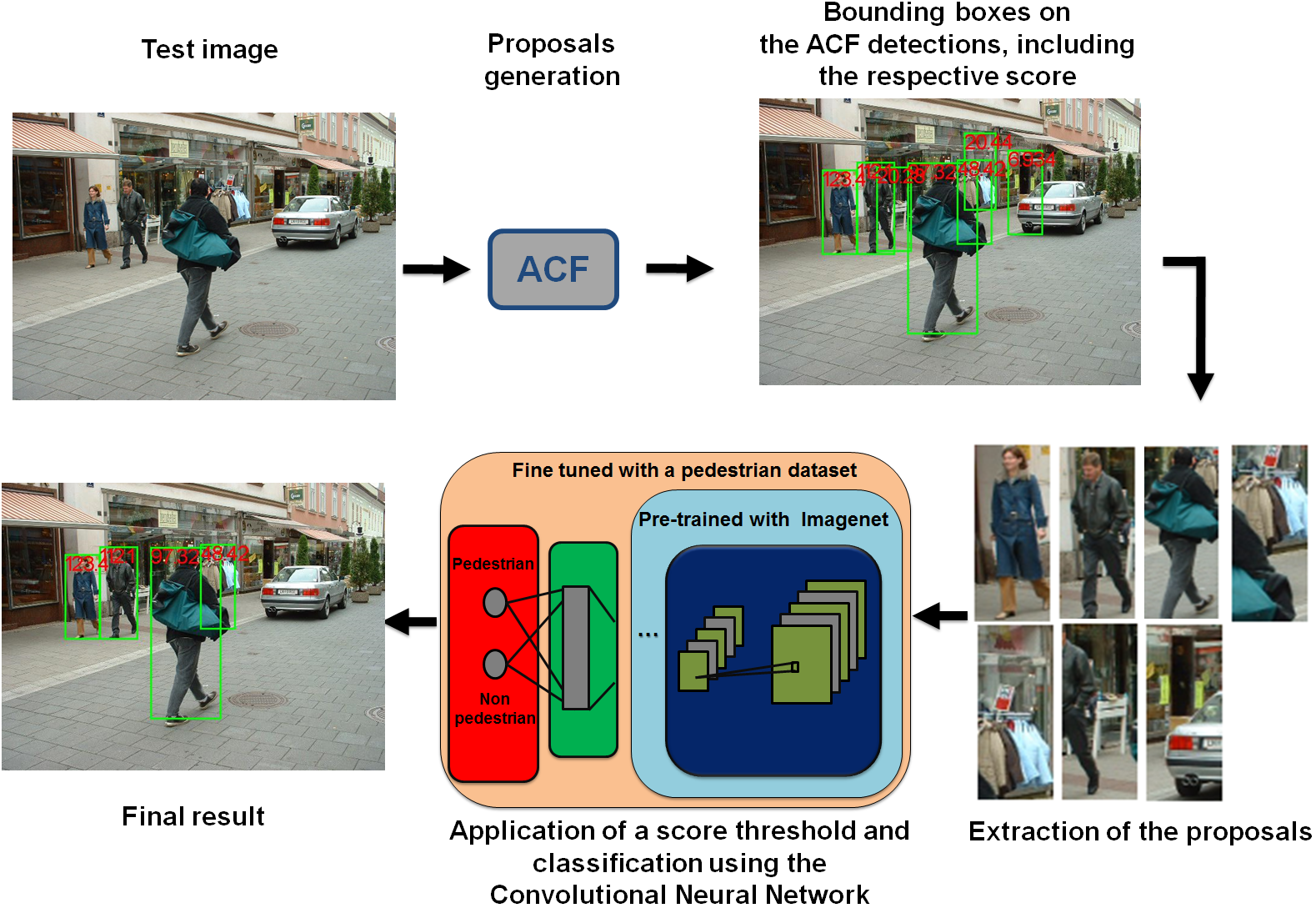}
  \end{center}
  \caption{Scheme of the proposed methodology. The Aggregate Channel Features (ACF) non-deep detector is cascaded with a deep Convolutional Neural Network (CNN). First, the ACF detector generates proposals by providing regions of interest that might contain pedestrians (see green rectangles in the third image). Then, these proposals are classified by the CNN (using the RGB feature map) to improve the accuracy and reduce the number of false positives.}\label{fig:PD-proposal}
\end{figure}

In this paper, we propose a solution for real-time PD using Computer Vision (onboard and/or offboard cameras) for people state estimation, using a novel deep learning technique. The scheme of our approach is shown in Fig.~\ref{fig:PD-proposal}. The solution was firstly tested using offboard and onboard images taken from our testbed and robot platform. Then, to validate our approach, we applied the proposed solution to a HAN problem. The results show that the proposed solution fulfills the respective goals.

The PD task is an important component of our framework, not only in terms of accuracy but also regarding speed, since real-time performance is required. The literature concerning the PD field of study is vast and has been evolving in order to provide improved solutions to this problem (surveys can be found in \cite{DollarPAMI2012,BenensonECCV2014}).
In general, to perform PD, a detection window is ``slided'' in several image locations separated by a certain stride, using multiple scales. Features are extracted and classified to determine the presence of a pedestrian. Finally, redundant detections are eliminated using non-maximal-suppression.
Initially, the PD methods employed features designed in a handcrafted fashion (e.g., Haar \cite{Viola2004}, including its informed version \cite{Zhang2014}). Nevertheless, the recent success of Convolutional Neural Networks (i.e. CNNs), achieved in several applications, such as, classification, localization and detection \cite{RussakovskyIJCV2015}, led to the adoption of this methodology to the PD task.

The computations associated with the CNN are expensive when compared to the ones required by methods using handcrafted features. Therefore, to improve the detector's speed, an hybrid solution can be adopted by cascading a faster and shallower method, based on handcrafted features, with a deep CNN. The handcrafted based approach generates proposals (i.e., promising regions for the pedestrians locations), whose classification is refined by the CNN (i.e., the accuracy is enhanced by removing false positives).

Furthermore, the transfer learning technique \cite{YosinskiNIPS2014} should be employed when training the CNN in order to prevent overfitting. This technique consists in transferring parameters from a network trained on an auxiliary task (with a large auxiliary dataset, e.g. Imagenet \cite{RussakovskyIJCV2015}) to initialize our model. Then, our initialized network is fine-tuned (i.e., retrained) for the task of interest (in our case, using the PD dataset).

This paper is organized as follows. Sec.~\ref{sec:Methodology-for-PD} describes the PD methodology, whereas Sec.~\ref{sec:Tracking} addresses the tracking procedure.  Section~\ref{sec:deep-trn-tst} describes how the CNN training is accomplished, and Sec.~\ref{sec:Adaptation of the CNN} mentions how a pre-trained model is used for fine-tuning. The PD performance is evaluated in the ``corridor'' and ``Mbot'' real scenarios (Section ~\ref{sec:Results-CNN-Corridor-Mbot}). Sec.~\ref{sec:results_real} presents the results of the overall and complete framework (PD + HAN). Finally, Sec.~\ref{sec:conclusions} draws conclusions and discusses further work.


\section{Vision-Based People Detection and Tracking through Deep Learning}
\label{sec:localization}

Regarding PD, we use a combination of handcrafted methods with deep learning methodologies. More specifically, first, the non-deep detector Aggregate Channel Features (ACF) \cite{DollarPAMI2014} provides regions of interest (i.e., proposals), allowing to reduce the expensive computational effort that would be required by a CNN, in the exhaustive process of sliding window search. Then, the proposals obtained previously are classified by the CNN, allowing to improve the accuracy of the ACF detections (as depicted in Fig.~\ref{fig:PD-proposal})\footnote{Other non-deep detectors could be used, but we adopted the ACF detector because it is fast.}.

\subsection{Methodology for PD}
\label{sec:Methodology-for-PD}

In the following, we describe the methodology and introduce the notation for the PD task. Given a training set ${\cal D}=\{ ({\bf x},{\bf y})_i\}_{i=1}^{|\cal D|}$, ${\bf x}$ represents the input image with ${\bf x}:\Omega\rightarrow\mathbb{R}^3$ and $\Omega$ representing the image lattice\footnote{In this paper, the RGB feature map is considered for the image ${\bf x}$.} of size $w \times h\times d$, with $d=3$; the class label is defined in ${\bf y}\in{\cal Y}=\{0,1\}^{C}$ that denotes the (absence) presence of the pedestrian in the $i$th image ${\bf x}_i$ (i.e., $C=2$). A detector (e.g., ACF or LDCF \cite{NamNIPS2014}) is applied to each input image ${\bf x}_i$ in order to generate proposals (including the associated confidence scores). This results in an output set denoted by ${\cal O}=\{ ({\bf x}({\cal B}), {\cal S})_i\}_{i=1}^{|\cal O|}$, where ${\cal B} = \{b_k\}_{k=1}^{|{\cal B}|}$
denotes the set of bounding boxes coordinates, with $b_k=[x_k,y_k,w_k,h_k]\in\mathbb{R}^{4}$ representing the top-left point and width and height enclosing (or not) the pedestrian. The content (i.e., the proposals) of the image delimited by the bounding boxes ${\cal B}$ is represented by ${\bf x}({\cal B})$, and ${\cal S}=\{s_k\}_{k=1}^{|{\cal S}|}$ correspond to the ACF detector confidence scores assigned to the proposals ${\bf x}({\cal B})$.

The use of pre-trained models, during the CNN initialization process, allows to obtain gains concerning the generalization ability of the model \cite{YosinskiNIPS2014}.
Hence, we select the VGG CNN model \cite{SimonyanICLR2015},
pre-trained with Imagenet \cite{RussakovskyIJCV2015}. We denote the dataset to pre-train the CNN as $\widetilde{\cal D}=\{ (\widetilde{\bf x},\widetilde{\bf y})_n \}_{n=1}^{|{\widetilde{\cal D}}|}$, with $\widetilde{\bf x}:\Omega\rightarrow\mathbb{R}^3$ and $\widetilde{\bf y}\in\widetilde{\cal Y}=\{0,1\}^{\widetilde C}$, where $\widetilde C$ is the number of classes in the pre-trained model (for the Imagenet, we have $\widetilde{C}=1000$).


\subsubsection{CNN model}
\label{sec:CNN-model}

Typically, the structure of CNNs includes the composition of: convolutional layers with a non-linear activation function; non-linear subsampling layers; fully connected layers; and a multinomial logistic regression layer \cite{KrizhevskyNIPS2012}.
Formally, the CNN can be denoted by:
\begin{equation}
  f({\bf v}, \theta^{(1)}) = {\bf v}^{\star}=f_{\rm out}\circ f_{L}\circ ...\circ f_2 \circ f_1({\bf v}^{(0)}),\label{eq:CNN1}
\end{equation}
where ${\bf v}^{(0)}={\bf v}$ is the input data, $\circ$ denotes the composition operator, $\theta^{(1)}$ represents the CNN parameters (i.e., the weights and biases), and ${\bf v}^{\star}$ is the CNN output (prediction). 
For the PD case, ${\bf v}^{(0)}={\bf v}={{\bf x}(\cal B}^{(0)})={{\bf x}(\cal B})$ (see $\text{4}^{\text{th}}$ image in Fig. \ref{fig:PD-proposal}), which represents the proposals, and ${\bf v}^{\star}= f({\bf v}, \theta^{(1)}) = {\bf y}^{\star}=f({\bf x}({\cal B}),\theta^{(1)})$, which denotes the prediction. The CNN is applied to these proposals, outputting the probability of the existence of a pedestrian in each one of them. If a proposal is classified as pedestrian, it is saved and no changes are made to its original ACF score. The proposals considered to be non pedestrians are eliminated, in order to reduce the number of false positives.

The convolution of a layer's input with a set of filters, followed by a non-linearity, is represented by:
\begin{equation}
  {\bf v}^{(k)} = f_k({\bf v}^{(k-1)}) = \sigma ({\bf W}_k(i,j)^{\top} {\bf v}^{(k-1)}+\beta_k),\label{eq:CNN2}
\end{equation}
where the convolutional filters are represented by the weight matrix ${\bf W}_k$ and the bias vector $\beta_k$, and where $\sigma(.)$ represents the non-linearity (e.g. the Rectified Linear Unit \cite{KrizhevskyNIPS2012}). The non-linear subsampling layers are denoted by ${\bf v}^{(k)}= \downarrow {\bf v}^{(k-1)}$, where $\downarrow$ represents the function (e.g., mean or max) applied to the input regions, leading to the size reduction. The fully connected layers employ a special case of the convolution represented in (\ref{eq:CNN2}), because the entire input is convolved with individual filters. The multinomial logistic regression layer uses the soft-max function: ${\bf y}(i)=\frac{\exp({{\bf v}^L}(i))}{\sum_j \exp({{\bf v}^L}(j))}$ to calculate the probability for each class (indexed by $i$), using the input ${\bf v}^L$ from the $L^{\text{th}}$ layer.

The loss function used during training is the binary cross entropy loss expressed as:
\begin{equation}
  {\cal L} = \frac{1}{|{\cal D}|}\sum_{i=1}^{|\cal D|} - y(i)\times \log(y^{\star}(i)) - (1-y(i)) \times \log(1-y^{\star}(i))\label{eq:binary-loss}
\end{equation}
where the training set ${\cal D}$ is indexed by $i$.

We denote a pre-trained CNN model as: $\widetilde{\bf y}=f(\widetilde{\bf x},\widetilde{\theta})$, with $\widetilde\theta=[\widetilde\theta_{\rm cn},\widetilde\theta_{\rm fc},\widetilde\theta_{\rm lr}]$, where $\widetilde{\theta}_{\rm cn}$ are the parameters for the convolutional and non-linear subsampling layers, $\widetilde{\theta}_{\rm fc}$ are the parameters for the fully connected layers, and $\widetilde{\theta}_{\rm lr}$ are the parameters for the multinomial logistic regression layer.

The parameters $\widetilde{\theta}_{\rm cn}$ and $\widetilde{\theta}_{\rm fc}$ (or a subset of them) can be transferred to another CNN model, in order to provide a rich initialization \cite{YosinskiNIPS2014}. The layers, whose parameters were not transferred, can be randomly initialized. Finally, the resulting CNN is fine-tuned with the dataset corresponding to the task of interest.

In the PD case, we transfer the parameters from the convolutional and non-linear subsampling layers, and randomly initialize all the other layers. Due to changes in the CNN input size, the parameters for the fully connected layers were adjusted accordingly. The multinomial logistic regression layer was adapted to consider only two classes (pedestrian and non-pedestrian). Finally, this CNN model for PD was fine-tuned with the pedestrian dataset ${\cal D}$, resorting to the binary cross entropy loss in (\ref{eq:binary-loss}).

\subsection{Tracking}
\label{sec:Tracking}

Taking the position measurements from the people detection scheme (described in the previous subsection), the goal of the tracking phase is to associate detections between frames and to estimate the direction of a person's velocity. For that purpose, we use a simple Kalman filter\footnote{Other filters could be used, but we used the Kalman filter because of its simplicity--this is not the main focus of the paper.}. Moreover, since humans tend to walk at a constant velocity \cite{35}, a constant velocity model was assumed. For each iteration, the PD detections are associated with the current tracks. This is performed using either the Nearest Neighbor or the Nearest Neighbor Joint Probabilistic Data Association methods. The associated detections are then used in the update phase of the respective track. If the detection is not associated with any track, it is stored, and if it is stable for some time, a new track is started. Finally, if a track has no detection associated with it for a while, it is deleted. For more details on the tracker and on the association methods used, please refer to \cite{bellotto:2010} and \cite{barshalom:2009}, respectively.


\section{Material and methods}\label{sec:Material-and-methods}
In this section, we describe the training details used for the pedestrian detector, taking into account the overall framework for the navigation setup.

\subsection{Dataset for the CNN training}\label{sec:deep-trn-tst}
The pedestrian dataset chosen to train the CNN was the INRIA dataset \cite{DalalCVPR2005}, which is a popular benchmark in the PD field of study\footnote{More details can be found at: \url{http://pascal.inrialpes.fr/data/human/}.}.
Originally, this dataset is divided in train (1832 images) and test (288 images) sets. Within the train set, there are 1218 negative images (i.e., without pedestrians), and 614 positive images (i.e., with pedestrians).

For the process of training the CNN, we extracted proposals from the original train set as follows. To construct the positive set, first, we use the ground truth positive training bounding boxes to extract proposals, resulting in ${\cal B}_{\rm pos}=1237$ samples. Then, we augment (i.e., with data augmentation) this set using two steps:
\begin{enumerate}
  \item{Horizontal flipping applied to ${\cal B}_{\rm pos}$, resulting in ${\cal B}^{(1)}_{\rm pos}=2474$ (including also
    ${\cal B}_{\rm pos}$); and}
  \item{Random deformations (by affecting pixels in the range ${\cal R}=[0,5]$ for the beginning and end) performed in the previous set ${\cal B}^{(1)}_{\rm pos}$, resulting in ${\cal B}^{(2)}_{\rm pos}=4948$.}
\end{enumerate}

To construct the negative set ${\cal B}_{\rm neg}$, we use the methodology from \cite{RibeiroPR2016} (i.e., applying a non-fully trained LDCF detector) to extract proposals from the negative images. This results in ${\cal B}_{\rm neg}=12552$ negative proposals.
The final set of CNN train proposals comprises a total of 17500 samples, which are divided in train (15751 proposals, i.e. 90$\%$ of the total) and validation (1749 proposals, i.e. 10$\%$ of the total).

\begin{figure*}
  \vspace{-0.25cm}
  \centering
  \subfloat[Pedestrian detection in the ``corridor'' sequence.]
  {\includegraphics[width=0.32\textwidth]{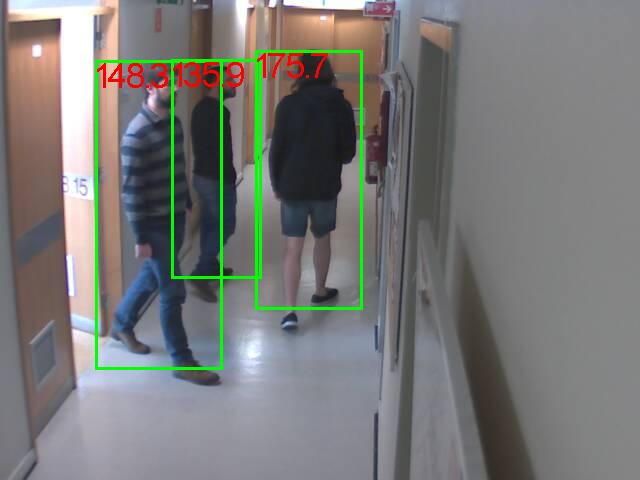}
  \includegraphics[width=0.32\textwidth]{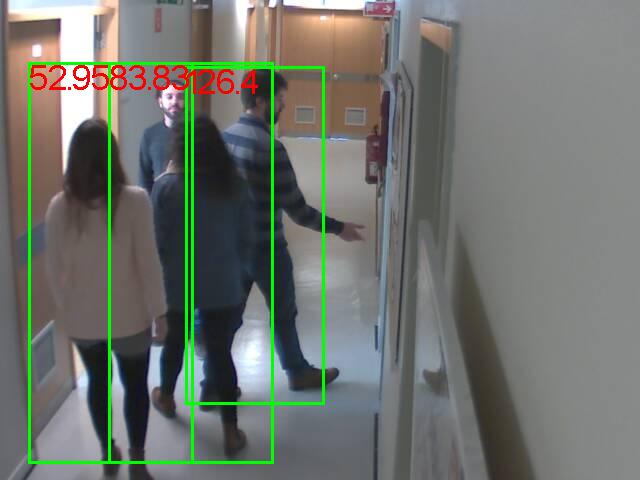}
  \includegraphics[width=0.32\textwidth]{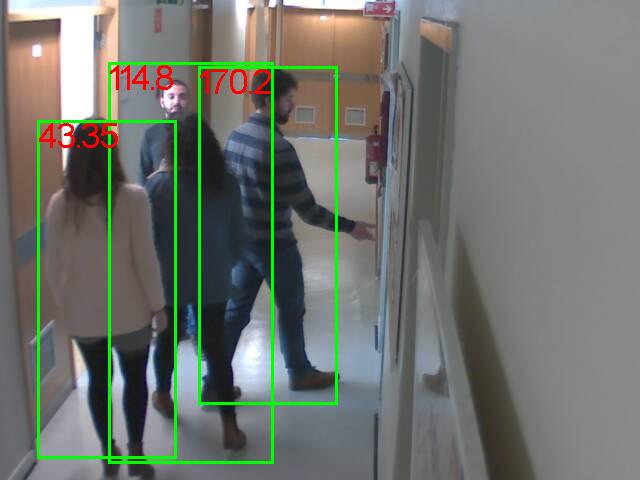} \label{fig:david1}}\\
  \subfloat[Pedestrian detection in the ``Mbot'' sequence.]
  {\includegraphics[width=0.32\textwidth]{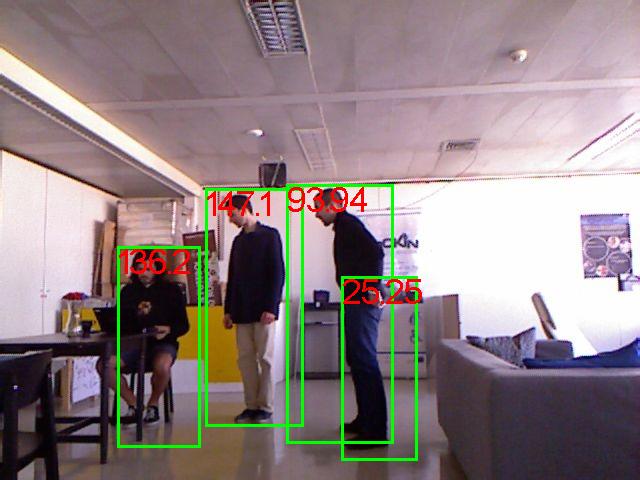}
  \includegraphics[width=0.32\textwidth]{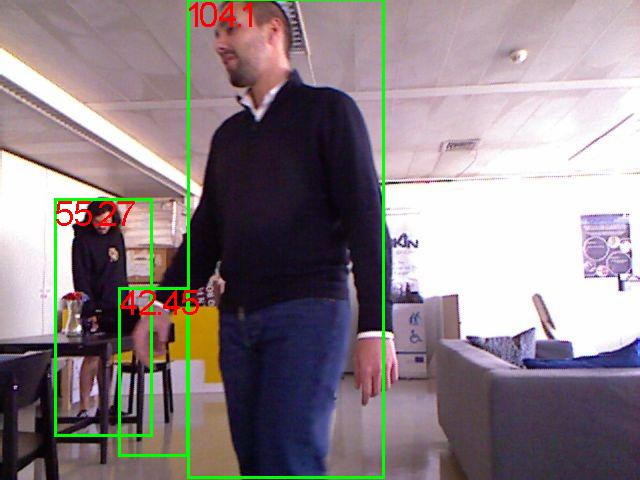}
  \includegraphics[width=0.32\textwidth]{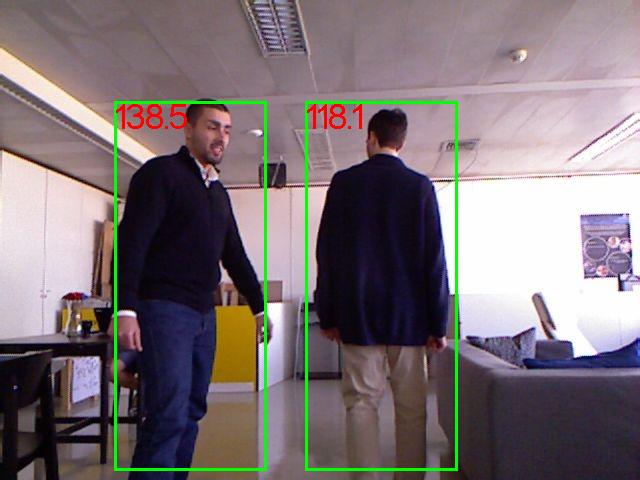}\label{fig:david2}}\\
  \caption{To test our PD methodology, we used two sequences of images acquired from possible real scenarios camera locations, which can be the ceiling and on the robot. In Figs.~\protect\subref{fig:david1} and~\protect\subref{fig:david2}, three images of the ``corridor'' and ``Mbot`'' datasets are depicted with the obtained detections (in green) and the corresponding scores (in red), respectively.}
  \label{fig:david-results}
\end{figure*}

\subsection{CNN model for training}\label{sec:Adaptation of the CNN}
Since we use a pre-trained CNN model (which can be considered a regularization technique), we have to adapt it to our task of interest (i.e., PD). Next, we mention the selected pre-trained CNN, the changes made, and the final architecture training details.

\subsubsection{Pre-trained model original architecture}
The VGG Very Deep 16 architecture (VGG-VD16) (configuration D) \cite{SimonyanICLR2015}\footnote{Additional details can be found at: \url{http://www.robots.ox.ac.uk/~vgg/research/very_deep/}.} was chosen to be the pre-trained architecture. This model's original input size is $224\times 224\times 3$, and has 13 convolutional layers (with a $3 \times 3$ window), the Rectified Linear Unit non-linearity, five max-pooling operations (with a $2 \times 2$ window with 2-times reduction), three fully connected layers and a multinomial logistic regression layer (see Sec. \ref{sec:CNN-model}). The classification results from the output of the last layer, which has 1000 filters corresponding to each one of the classes in ILSVRC \cite{RussakovskyIJCV2015}.
The dataset used to perform the pre-training of this model is Imagenet \cite{RussakovskyIJCV2015}, containing 1K visual classes, 1.2M training, 50K validation and 100K test images.

\subsubsection{Pre-trained model changes and fine-tuning}
Motivated by the pre-trained model's expensive and time consuming computations, the original dimensions of the CNN input were downscaled from $224 \times 224 \times 3$ to $64 \times 64 \times 3$. With this modification, inference cannot be performed after the first fully connected layer. Furthermore, the classification related layer must be adapted to transition from 1000 ILSVRC classes to two PD classes (i.e., pedestrian and non-pedestrian). To overcome this problems, we randomly initialize the parameters of the three fully connected layers with the correct dimensions. For this initialization procedure, we selected a Gaussian distribution, with mean ${\cal \mu} =0$ and variance $\sigma^2 =0.01$. The modified CNN model is fine-tuned with the positive and negative proposals training sets, acquired from the INRIA dataset (as described in Sec. \ref{sec:deep-trn-tst}).

In terms of the fine-tuning hyperparameters, we used 10 epochs with a minibatch of 100 samples, a learning rate of 0.001, and a momentum of 0.9.

For the test, first, the proposals (i.e., promising regions for the existence of pedestrians) are extracted by running the ACF detector in the test images. Then, these proposals are classified as pedestrians or non-pedestrians, by the fine-tuned CNN model described previously.

\subsubsection{Implementation}
The PD methodology was implemented in MATLAB, running on CPU mode on 2.50 GHz Intel Core i7-4710 HQ with 12 GB of RAM and 64 bit architecture. To run the ACF detector and to evaluate the performance, the Piotr's Computer Vision MATLAB Toolbox \cite{DollarToolbox} (2014, version 3.40) was employed. Concerning the CNN framework, we utilized the MatConvNet toolbox \cite{vedaldi15matconvnet}. The experiments described in the next section, namely: Sec. \ref{sec:Experimental-results}, and Table \ref{tab:running-times-2}), were conducted using the same settings.

\section{Experimental Results}\label{sec:Experimental-results}
In this section, we evaluate the PD method in two real scenarios (i.e., datasets named "corridor" and "Mbot"). To conclude, we validate our approach by using the PD method on a Human-Aware Navigation problem, in which a real-time detection performance is required.

\subsection{Evaluation of the PD method}\label{sec:Results-CNN-Corridor-Mbot}
In order to conduct experiments in real scenarios, we acquired two indoor datasets and tested the proposed PD method on them. These datasets are: 1) the "corridor" dataset, which comprises 5556 images, and 2) the "Mbot" dataset, which comprises 3966 images. For both sets, the image (i.e., frame) dimensions are $480\times 640$. The detection results for some samples of these datasets are depicted in Fig.~\ref{fig:david-results}.

For each dataset, we measure the runtime of the final PD method (i.e., ACF+CNN) proposed in Sec.~\ref{sec:Adaptation of the CNN}. As a result, we obtain approximately 707.27 seconds, which is equivalent to 7.85 FPS, for the "corridor" dataset (5556 frames), and 839 seconds, which is equivalent to 4.84 FPS, for the "Mbot" dataset (3966 frames). Further details about the runtime figures are presented in Tab.~\ref{tab:running-times-2} (top, the two columns in the field named "Baseline"), where the values represent per frame metrics.

To reach the real-time specifications required in robot navigation tasks, the speed should be improved. This can be accomplished by filtering the ACF proposals based on the confidence score, since this procedure reduces the number of proposals to be processed by the CNN, increasing the detection speed. The goal is to improve the previous speed, and achieve real-time performance, without substantially degrading the accuracy.

Taking into account that the confidence scores are important indicators to determine the relevance of each proposal, a score rejection threshold can be established. Only the proposals with score above this threshold are classified by the CNN. Following \cite{VermaWICCV2015}, we selected a threshold value of 40.

Consequently, in the process of discarding false positives, first we should eliminate the easier ones resorting to the threshold operation, and then we should eliminate the harder ones using the CNN. The possible loss in accuracy versus the speed improvement is determined by the choice of the threshold value.

Accordingly, using the threshold operation strategy, the detection speed of the overall method (i.e., ACF+CNN) is improved, in comparison with the baseline metrics. The cases before (``Baseline'' field) and after (``Threshold'' field) the threshold operation are depicted in Tab.~\ref{tab:running-times-2}. As presented in Tab.~\ref{tab:running-times-2}, we are able to achieve the real-time requirements for our navigation setup, by reaching a detection speed of approximately 10 FPS.

\begin{table}
  \caption{Runtime figures before (top, ``Baseline'') and after (bottom, ``Threshold'') the threshold operation applied to the ACF proposals, when using the overall PD method (i.e., ACF+CNN).}
  \label{tab:running-times-2}
  \begin{center}
    \begin{tabular}{|c|c|c|}
      \hline
      {\bf  Dataset} & {\bf  Data seq. 1 (corridor)} & {\bf Data seq. 2 (Mbot)}   \\ \hline

      &              Total time       =  0.1273 sec.       &    Total time  =  0.2066 sec.       \\
      Baseline    &  ACF time         =  0.0326 sec.       &    ACF time    =  0.0367 sec.       \\
      &              CNN time         =  0.0947 sec.       &    CNN  time   =  0.17   sec.       \\
      &              {\bf Frame rate} =   {\bf 7.85} FPS   &    {\bf Frame rate} =   {\bf 4.84}  FPS \\
      \hline

      &              Total time  =  0.0961 sec.            &    Total time  =  0.1026 sec.       \\
      Threshold   &  ACF time    =  0.0333 sec.            &    ACF time    =  0.0381 sec.       \\
      &              CNN time    =  0.0628 sec.            &    CNN  time   =  0.0645 sec.       \\
      &              {\bf Frame rate} =   {\bf 10.41} FPS  &    {\bf Frame rate} =   {\bf 9.74} FPS  \\
      \hline
    \end{tabular}
  \end{center}
\end{table}


\subsection{Real Experiments in an Indoor Scenario}
\label{sec:results_real}

\begin{figure}
  \subfloat[]{\includegraphics[width=0.135\textwidth]{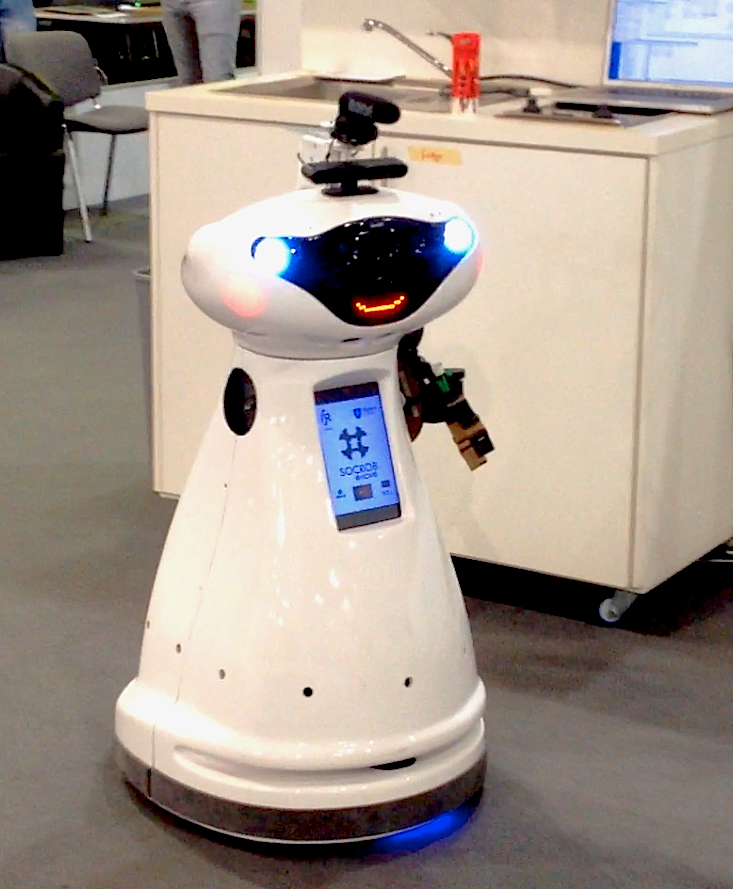}\label{fig:real_setup:mbot}}\,
  \subfloat[]{\includegraphics[width=0.21\textwidth]{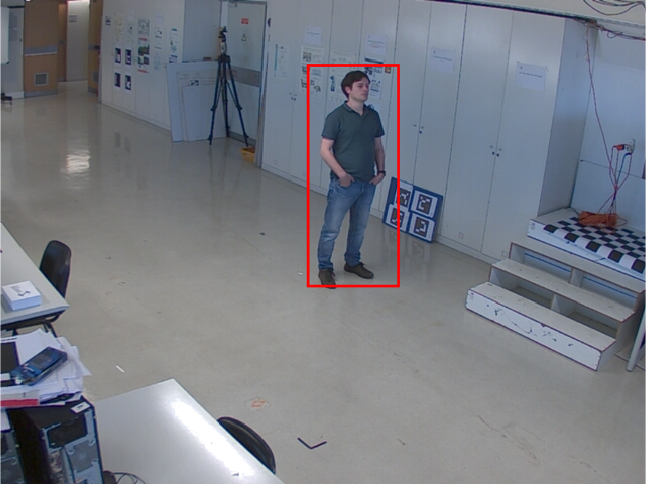}\label{fig:real_setup:camera2}} \,
  \subfloat[]{\includegraphics[width=0.117\textwidth]{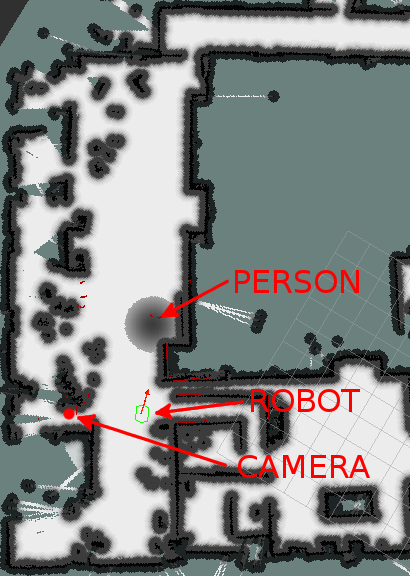}\label{fig:real_setup:rviz}}
  \caption{Representation of the setup used in the experiments. Fig.~\protect\subref{fig:real_setup:mbot} shows the robot platform and Fig.~\protect\subref{fig:real_setup:camera2} shows an image of the camera that will be used to estimate the pedestrians (as it can be seen, in this image we already show the bounding box identifying a person in the environment). To conclude, Fig.~\protect\subref{fig:real_setup:rviz} shows the environment (ROS rviz package), with the position of all the cameras, the position of the robot, and the pedestrian, with the respective HAN constraint (in this case the pedestrian was standing).}
  \label{fig:real_setup}
\end{figure}

In this section we evaluate our PD framework on a Human-Aware Navigation (HAN) application. For that purpose, we consider a setup as shown in Fig.~\ref{fig:real_setup}:
\begin{itemize}
  \item{We use a MBOT mobile platform \cite{72} (see Fig.~\ref{fig:real_setup}\subref{fig:real_setup:mbot}) in a typical domestic indoor scenario; and}
  \item{A perspective camera mounted on the ceiling (an example of an image acquired using this camera is shown in Fig.~\ref{fig:real_setup}\subref{fig:real_setup:camera2}) was placed in the environment as shown in Fig.~\ref{fig:real_setup}\subref{fig:real_setup:rviz}.}
\end{itemize}

The HAN is not the focus of the paper. Then, we follow the navigation (including the constraints associated with the HAN) proposed at \cite{Mateus:2015}. Basically, the authors use the $A^*$ as a path planner (to ensure a minimal cost path) and define a set of HAN constraints as cost functions. These cost functions are shown in the figures of the experimental results (Figs. 4 (a) and (b)), and are computed using the following procedure:
\begin{enumerate}
  \item{Selecting the middle point of the lower edge of the bounding box that is given by the PD;}
  \item{Projecting this middle point on the image onto the floor plane (assuming that the position of the robot is known); and}
  \item{Estimating the pedestrian velocity in the world coordinate system.}
\end{enumerate}

\begin{figure*}
  \vspace{-0.25cm}
  \centering
  \subfloat[Experiment 1, where people are standing during all the robot's motion.]{
    \shortstack{\includegraphics[width=0.193\textwidth]{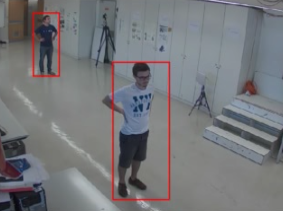}
    \includegraphics[width=0.193\textwidth]{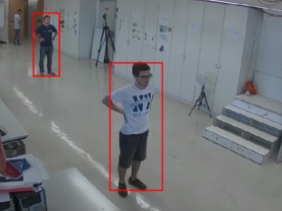}
    \includegraphics[width=0.193\textwidth]{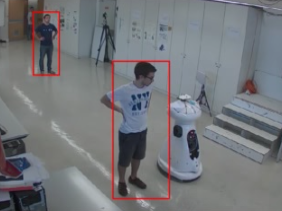}
    \includegraphics[width=0.193\textwidth]{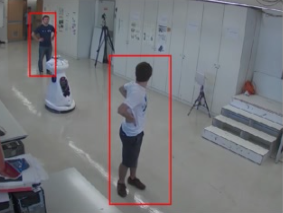}
    \includegraphics[width=0.193\textwidth]{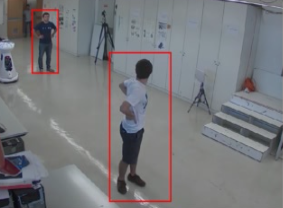}\\
    \includegraphics[width=0.193\textwidth]{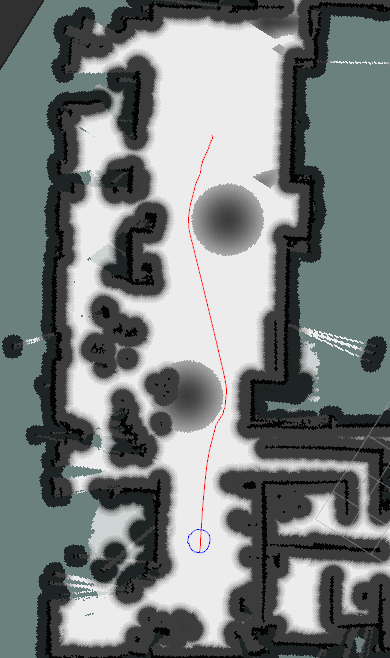}
    \includegraphics[width=0.193\textwidth]{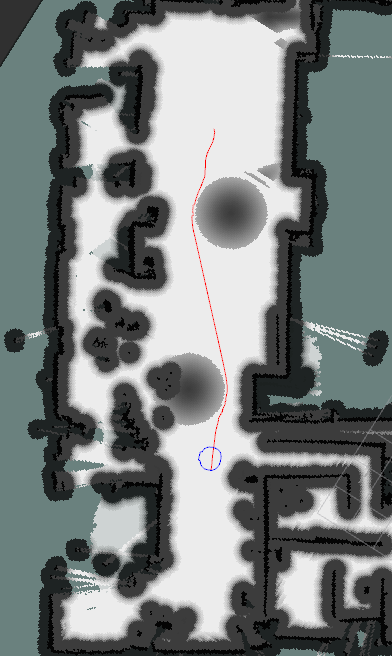}
    \includegraphics[width=0.193\textwidth]{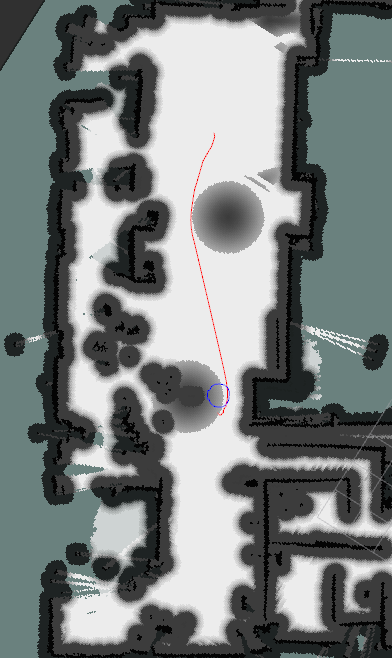}
    \includegraphics[width=0.193\textwidth]{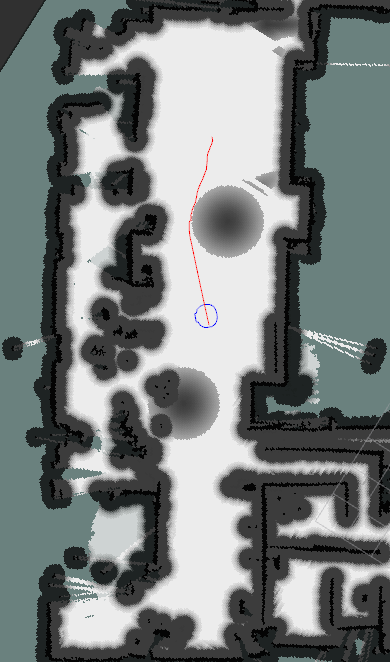}
    \includegraphics[width=0.193\textwidth]{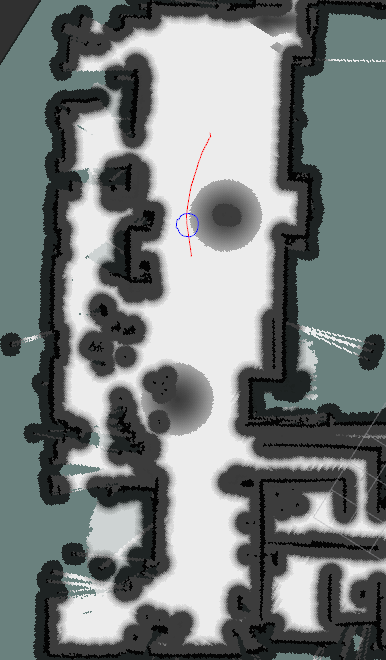}}\label{fig:experiment1}
  }\\
  \subfloat[Experiment 2, where people are standing in the beginning but, when the robot starts moving, one person will start walking in the robot's path.]{
    \shortstack{\includegraphics[width=0.193\textwidth]{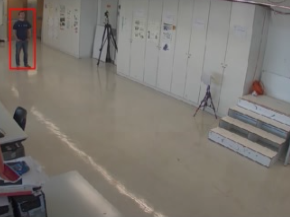}
    \includegraphics[width=0.193\textwidth]{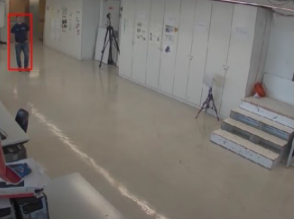}
    \includegraphics[width=0.193\textwidth]{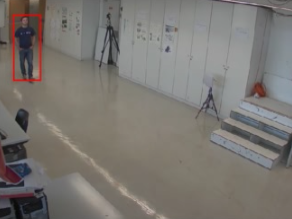}
    \includegraphics[width=0.193\textwidth]{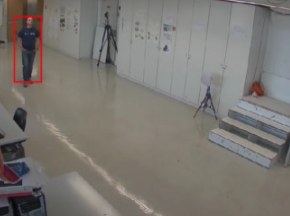}
    \includegraphics[width=0.193\textwidth]{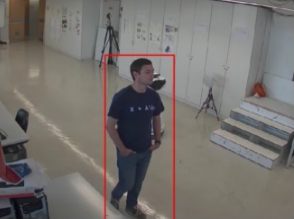}\\
    \includegraphics[width=0.193\textwidth]{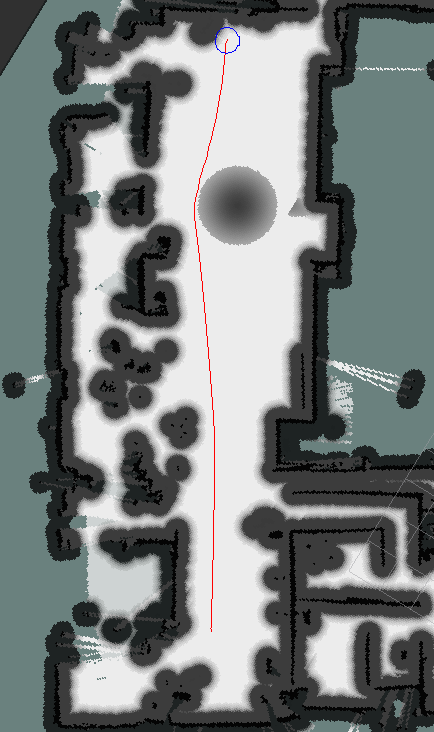}
    \includegraphics[width=0.193\textwidth]{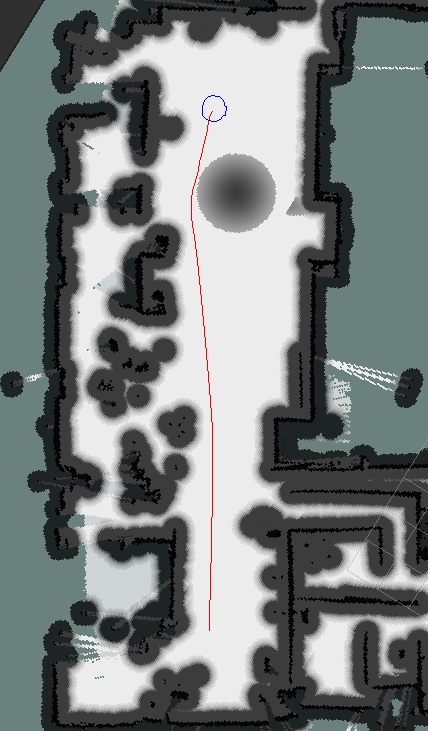}
    \includegraphics[width=0.193\textwidth]{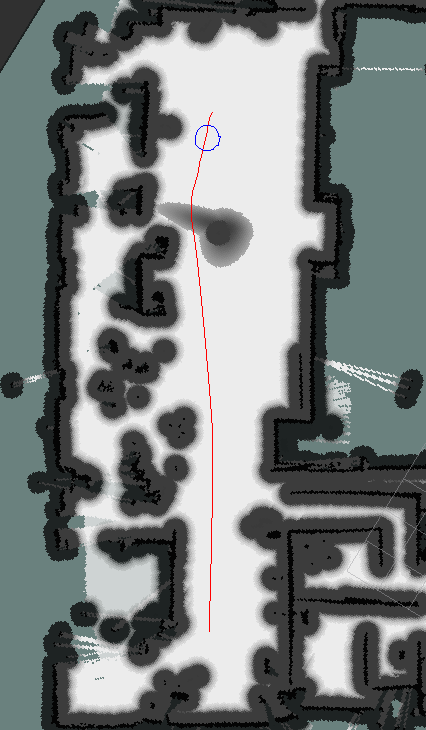}
    \includegraphics[width=0.193\textwidth]{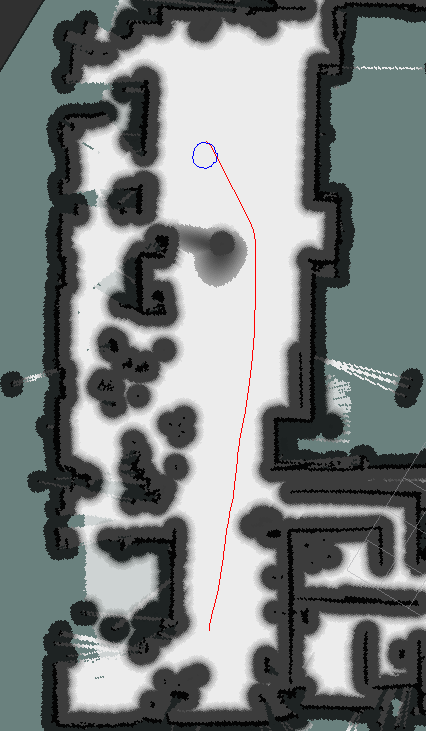}
    \includegraphics[width=0.193\textwidth]{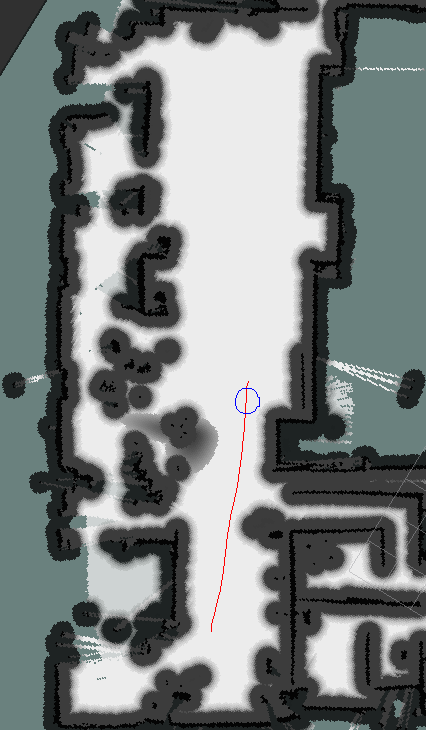}}\label{fig:experiment2}
  }
  \caption{Results of the real experiments in realistic scenarios. A MBOT mobile robot \cite{72} is used on a typical domestic indoor scenario, developed for benchmarking in an ERL@Home testbed. In these experiments, the robot is navigating while a person (which firstly was standing) starts walking, blocking the robot's path. At that moment, the robot must overtake the pedestrian according to the respective social rule, that says that a robot must overtake a person through the left. The robot is shown as a blue circle while the path of the robot is shown as a red line.}
  \label{fig:experiments_real_scene}
\end{figure*}

The goal of these experiments is to evaluate the proposed PD on the image, using a robot navigation application in the presence of people. Two experiments were conducted, in which we apply our PD and the previous method for HAN:
\begin{enumerate}
  \item{Firstly, we consider a simple example where a robot is going towards a goal and people are standing in the environment (in front of the robot). The robot must take their positions into account (which are given by the PD mapped onto the floor plane) on the path planning, in order to avoid a collision; and}
  \item{In the second experiment, a person starts walking when the robot is moving, blocking the path of the robot. Following the social rules, the robot must replan its path, to overtake the person by the left.}
\end{enumerate}
The results of both experiments are shown in Figs.~\ref{fig:experiments_real_scene}\subref{fig:experiment1} and \ref{fig:experiments_real_scene}\subref{fig:experiment2}, respectively. Videos with these experiments will be included in the authors websites. As it can be seen by these figures, the robot behaves as expected, which proves that our PD method is suitable for robot navigation tasks, in the presence of people.


\section{Conclusions}
\label{sec:conclusions}

This paper presents a novel framework that integrates pedestrian detection in the problem of robot navigation. More specifically, it integrates a novel pedestrian detection approach jointly with specific motion constraints, representing the human-aware concerns. The novelty inherent to the PD methodology, is that it allows to improve the accuracy of a non-deep detector, by efficiently cascading a CNN. The PD method is evaluated using two sets of real images acquired on a typical robot navigation environment (considering both on-board and external camera sensors). The results show that the proposed solution is suitable for robot navigation tasks, namely in terms of both runtime and robustness. In addition, two experiments were conducted in a realistic scenario to assess the overall framework performance.

As future work, we are planning on fusing data from multiple external and on-board cameras, and other types of sensors, such as lasers, as well as test the proposed framework with different people behaviors.

\section*{Acknowledgements}
This work was partially supported by FCT[UID/EEA/50009/2013], and by the FCT grant SFRH/BPD/111495/2015.

We would also like to thank to Luis Luz for his help getting the experimental results.

\bibliographystyle{IEEEtran}
\bibliography{root}



\end{document}